\documentclass[conference]{IEEEtran}
\usepackage{amsmath,amsfonts}
\usepackage{algorithm}
\usepackage{algorithmic}
\usepackage{array}
\usepackage{textcomp}
\usepackage{stfloats}
\usepackage{url}
\usepackage{graphicx}
\usepackage{cite}
\usepackage{hyperref}
\usepackage{cite}
\usepackage{xcolor}
\def\BibTeX{{\rm B\kern-.05em{\sc i\kern-.025em b}\kern-.08em
    T\kern-.1667em\lower.7ex\hbox{E}\kern-.125emX}}

\begin{document}

\title{TRAX: TRacking Axles for Accurate Axle Count Estimation}

\thanks{This paper was produced by the AI Vision @ OCI. They are in Bengaluru, In.}


\author{
Avinash Rai, Sandeep Jana, Vishal Vijay \\
}

\maketitle

\begin{abstract}
Accurate counting of vehicle axles is essential for traffic control, toll collection, and infrastructure development. We present an end-to-end, video-based pipeline for axle counting that tackles limitations of previous works in dense environments. Our system leverages a combination of YOLO-OBB to detect and categorize vehicles, and YOLO to detect tires. Detected tires are intelligently associated to their respective parent vehicles, enabling accurate axle prediction even in complex scenarios. However, there are a few challenges in detection when it comes to scenarios with longer and occluded vehicles. We mitigate vehicular occlusions and partial detections for longer vehicles by proposing a novel TRAX (Tire and Axle Tracking) Algorithm to successfully track axle-related features between frames. Our method stands out by significantly reducing false positives and improving the accuracy of axle-counting for long vehicles, demonstrating strong robustness in real-world traffic videos. This work represents a significant step toward scalable, AI-driven axle counting systems, paving the way for machine vision to replace legacy roadside infrastructure.
\end{abstract}


\section{Introduction}
\IEEEPARstart {P}recise axle counting is the backbone of contemporary traffic management, tolls, and road planning. Axle count is needed to determine vehicle class, estimate road use, enforce weight limits, and calculate toll fees. Conventional means of axle count, which include pressure sensors, inductive loops, and visual counts, suffer from deficiencies in being scalable in size, installation-intensive, and reliable in various environments. As intelligent transport systems (ITS) become widespread in the road network, the need for automated, in-real-time and reliable axle counts has increased dramatically.

Computer vision and machine learning breakthroughs in the form of deep learning models such as YOLO (You Only Look Once)\cite{ref7} and DeepSORT (Deep Simple Online Realtime Tracking)\cite{ref5} have dramatically improved object detection and tracking capabilities. Building on these breakthroughs, video based car inspection has evolved into a promising technology in traffic analytics with cost effective solutions that can be scaled up. Axle counting in video images proves to be a challenging task in realistic scenes may be occluded partially, overlap with other objects, or extend over multiple video frames because of length and speed. Such complications need advanced algorithms to allow accurate detection, association, and counting of vehicle axles. This paper propose an automated pipeline that overcomes these challenges by combining state-of-the-art object detection methods and association algorithms that support real-time axle counting. This method is developed on the basis of the YOLO-OBB\cite{ref8} (Oriented Bounding Box) model to detect and classify trucks, semi-trucks, sedans, and trailers. A secondary model of YOLO\cite{ref8} is used to detect the tires, which are later associated with the corresponding vehicles. This hierarchical detection method allows precise axle counting of the individual vehicles and their attached trailers with ease in complicated video surveillance settings.

The originality of the system is in its resilience to partial occlusions and problematic cases of long cars spanning multiple frames. By employing tracking of tire detections and their positional relation with cars, the method provides reliable and precise axle count determination. In addition, the designed system is capable of working in real time and is thereby applicable to deployment in dynamic and high-traffic conditions of toll gates, highway monitoring, and intersections of cities.

We introduce a precise methodology as the pipeline of the proposed axle count and assess its performance against benchmark video datasets. Its performance is shown to be capable of providing accurate axle counts in conditions where other methods fail, which speaks to its capacity to improve traffic management systems to be more accurate and efficient. The proposed method seeks to overcome the gap between computer vision theory and deployment realities in transport systems and set the bar higher in video-based axle count solutions..

\section{Related Works}
Recent works investigated camera-based axle counting by identifying tires or axles in video. Compared to expensive in-road sensors, object detection supports real-time axle estimation using video. Miles et al. (2022) \cite{ref1} created a roadside system with two YOLOv3 networks, one for tire and one for vehicle detection, demonstrating 95\% mAP tire detection and 93\% accurate axle counting when all tires were seen. Kalman Filter tracking between frames was used to associate tire detections, though they observed decreased performance in cases of long-lasting occlusions. This indicates CNN-based tire detection competes with shape-based classifiers but has to manage occlusions and camera angles. In like manner, Souza et al. (2024) \cite{ref2} created an axle detection and counting system for free-flow tolling using various YOLO family models but found YOLOv5m to be best (99.40 precision, 98.20 recall). To account for long trucks only visible in multiple frames, they stitched frames together. Their solution identifies two main strategies: utilizing sophisticated YOLO versions to detect tires and multi-frame schemes to achieve full overlap in axle coverage.

\subsection{YOLO-based Vehicle and tire Detection}
Most methods employ YOLO-family detectors to detect either tires or entire axles from video frames. For example, Li et al. (2021) \cite{ref3} designed a higher-precision version of YOLOv5s\cite{ref8} to count axles and recognize tire type. Authors observe that side views of long trucks usually lack rear axles and so full-vehicle views become difficult to acquire. Instead of multi-frame stitching (slow), authors utilize a YOLOv5s detector and object tracking: detect vehicles first, tires in each vehicle secondly, and track them over time to estimate axle count. Essentially, tires in each frame in YOLOv5s get detected and a straightforward tracker consolidates them to axle counts. Similarly, Marcomini and Cunha (2022) compared three deep detectors (YOLO, Faster R-CNN, SSD) in the case of trucks' images. Authors state that YOLO (latest unspecified version) and SSD produced similarly great performance (around 96\% mAP) in detecting axles \cite{ref4}. Author's results indicate that single-shot detectors such as YOLO/SSD can achieve similar performances to two-stage detectors in detecting axles but at much faster speeds (they explicitly measured in FPS as a metric). Then, a straightforward count algorithm (tracker) allocates tires to vehicles and counts them to acquire axle count.

\subsection{Addressing Long Vehicles and Occlusions}
A major challenge is that long multi-axle vehicles usually do not fit entirely in a single frame. Li et al. \cite{ref3} comment that because of limited field of view, it is “hard to get the complete vehicle” in a single image; trucks often need to pass through multiple frames.
To accomplish this, some methods build up composite images (as in \cite{ref2} Souza et al.) or track the same car through frames (Li et al.\cite{ref3}, Miles et al.\cite{ref1}). The tolling method of Souza et al \cite{ref2}. takes the center slice of each consecutive frame to create a mosaic of the approaching vehicle so that all axles show up. Miles et al. \cite{ref1} track tires through frames with a Kalman filter and make explicit that long occlusions (say another vehicle passing in front) result in missed axle counts. They count a vehicle as partially occluded if some tires are occluded; in those instances their YOLOv3 detector would miss axles and thus accuracy decreases. In their experiments, the 93\% axle-count accuracy only applies “when all axles are visible” – implying lower accuracy in case of occlusion.

\section{System Architecture}

\begin{figure}[t] 
    \centering
    \includegraphics[width=\linewidth]{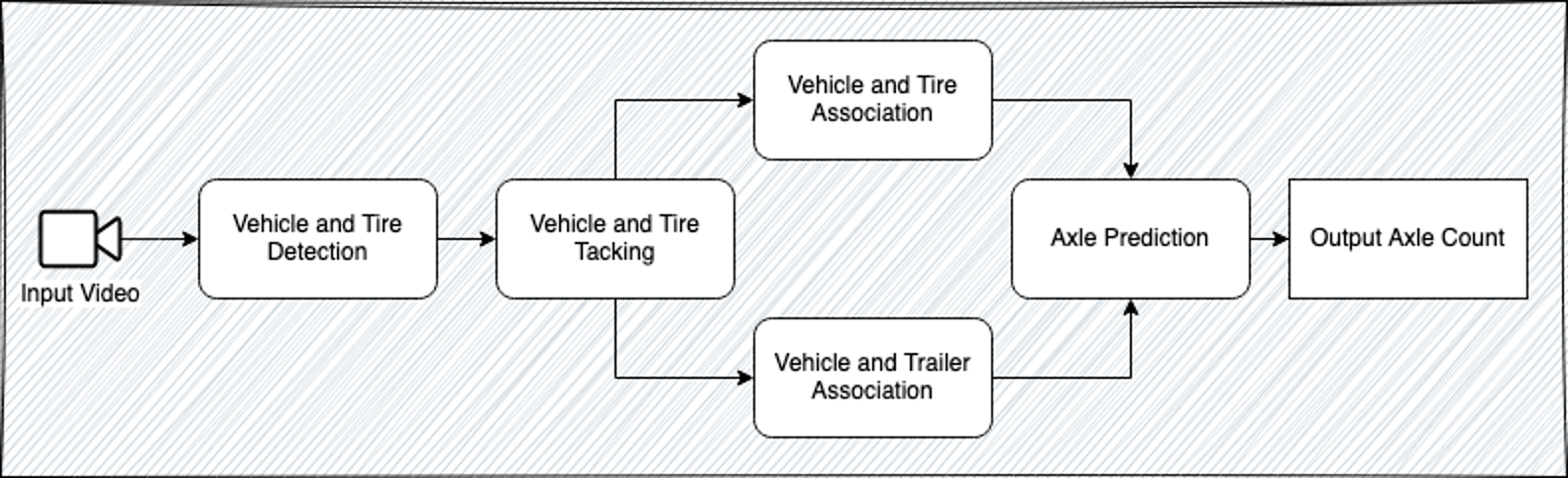} 
    \caption{System Architecture overview}
    \label{fig:pipeline-overview} 
\end{figure}

Figure \ref{fig:pipeline-overview} shows a high-level overview of the end-to-end system architecture. The pipeline comprises the following components: vehicle and tire detection, vehicle and tire tracking, vehicle and tire association, vehicle and trailer association, axle prediction as the final output.

The pipeline takes video $V$ as input, reads the video frame by frame, each frame $F$ is passed to vehicle detector and tire detector YOLOv5s\cite{ref8} models. The output of vehicle detector is a collection of detected $N$ vehicle boxes $V_{N}$ (OBB - (Oriented Bounding Boxes)) and the output of tire detector is a collection of $M$ detected tire boxes $Ti_{M}$ ( AABB -Axis-Aligned Bounding Boxes ).\\ 
\begin{enumerate}
    \item {Vehicle bounding boxes $V_{N}$ are then passed to vehicle $V$ and trailer $Tr$ association module to associate trailers to respective vehicles pulling them.}
    \item {Vehicle $V_{N}$ and tire boxes $Ti_{M}$ are passed as input to the vehicle and tire association module to associate tires to their respective vechiles and trailers.}
    \item {Once the tires and trailers are associated with vehicles. The vehicles and trailers with associated tires are passed to Axle counting module. Axle counting module then predicts the final axle count for each vehicle and trailers, also predicts the final axle count for vehicles which have associated trailers by combining their axle counts.}
\end{enumerate}

\subsection{Vehicle Detector}
\label{sebsec:vehicle_detector}
Axis-Aligned Bounding Boxes (AABB) in object detection form rectangular boxes aligned with axes and provide computational efficiency at the expense of possibly including additional background when objects are tilted. On the other hand, Oriented Bounding Boxes (OBB) adjust to accommodate an object's orientation, represent a tighter and hence superior fit and accuracy, particularly when objects tilt or take irregular shapes. Although OBB improves accuracy, they necessitate higher complexity and computation and are useful in cases of, e.g., aerial imagery where objects tend to appear at different angles. \\
In this research, we used YOLO-OBB\cite{ref8} (Oriented Bounding Boxes) as a vehicle detection model to improve precision in vehicle detection and classification. In contrast to traditional Axis-Aligned Bounding Boxes (AABB) that limit detection to rectangular areas aligned with image axes, OBB enables bounding boxes to be aligned with vehicle orientation. This adjustment offers a geometrical fit that is better in case of vehicles at skewed angles, eliminating as much irrelevant background as possible. Furthermore, OBB easily solves the overlap detection problem, such as multiple vehicles in an individual bounding box, especially in cases of densely crowded traffic or vehicles in neighboring lanes. With this improvement, detection accuracy is enhanced, and robustness of performance is improved in complicated real-world scenarios.

\subsection{Tire Detector}
The tire detection module uses YOLOv5s\cite{ref8}, designed to utilize axis-aligned bounding boxes with input size of 640x640 pixels. Tires, unlike vehicles, have round shapes and do not pose any particular angular orientations, and hence oriented bounding boxes are unnecessary. Tires are small objects relative to vehicles, and higher input resolution is required to extract minute details and provide accurate detection. The vehicle detection model YOLO-OBB\cite{ref8}, however, uses lower input resolution of 320x320 since larger objects like vehicles can be easily detectable even at low input sizes. The reason behind this is due to the nature of detection models like YOLO\cite{ref7}, where small objects make use of higher input resolution to address spatial details and large objects can still be detectable using mid to low input sizes and hence less computational overhead with minimal loss to detection performance. This custom design maintains optimal accuracy and efficiency in both tire as well as vehicle detection.

\subsection{Vehicle Tracker}
Vehicle tracking and real-world object tracking typically necessitate strong and precise algorithms to deal with complex environments. The tracking of vehicles and objects in real-world environments is facilitated in this work by using YOLO-OBB\cite{ref8} to detect vehicles and DeepSORT \cite{ref5} to track objects. The oriented bounding boxes give a necessary backbone to detect vehicles in dynamic environments and scenes with densely distributed or rotated objects. To keep objects identified in between frames, DeepSORT \cite{ref5}, a widely used multi-object tracking algorithm, is used. DeepSORT \cite{ref5} keeps vehicle tracking highly accurate by resolving challenges such as occlusions, heterogeneities in motion, and appearance variation of objects over time. OSNet\cite{ref6} is employed as ReID model for visual matching in DeepSORT.

\subsection{Vehicle and Tire Association}
Vehicle and Tire association is achieved using the Intersection over Union (IoU) metric, which quantifies the overlap between a detected tire bounding box and a vehicle bounding box. The IoU is mathematically defined as:

\[
\text{IoU} = \frac{\lvert B_{\text{tire}} \cap B_{\text{vehicle}} \rvert}{\lvert B_{\text{tire}} \cup B_{\text{vehicle}} \rvert}
\]

Here, \( B_{\text{tire}} \) and \( B_{\text{vehicle}} \) denote the bounding boxes of the tire and vehicle, respectively. The numerator \( \lvert B_{\text{tire}} \cap B_{\text{vehicle}} \rvert \) represents the area of their intersection, while the denominator \( \lvert B_{\text{tire}} \cup B_{\text{vehicle}} \rvert \) represents the area of their union. For each detected tire, the vehicle bounding box with the maximum IoU value is selected as the best match, provided the IoU exceeds a predefined threshold. This approach ensures accurate association of tires to vehicles based on the highest overlap, enabling precise tracking and analysis.

\subsection{Trailer and Vehicle Association}

Trailer and vehicle association is realized by adopting a systematic method of linking trailers and corresponding carrier vehicles with precision. The procedure starts by identifying objects of trailers and vehicles using the YOLO-OBB's provided class labels. After detection, a direction vector is computed for every detected trailer object, calculated by extracting its motion direction in the image frame. The direction vector $Dv$ is then prolonged in the same direction, either till it intersects with a vehicle object or goes off the image frame. If the extended vector intersects with any vehicle object, it is considered to be that respective trailer's carrier. The association technique is dependent on spatial overlap and motion path, making it strong enough to produce linkage even in environments with multiple trailers and vehicles. Elaborated in Algorithm \ref{algo:trailer-and-vehicle} below.

\begin{algorithm}
\caption{Trailer and Vehicle Association}
\label{algo:trailer-and-vehicle}
\begin{algorithmic}[1]
\STATE \textbf{Input:} Detected trailer objects, detected vehicle objects, class labels from YOLO-OBB
\STATE \textbf{Output:} Associated trailer-vehicle pairs
\FOR{each trailer object $t$}
    \IF{trailer object $t$ is detected}
        \STATE Calculate direction vector $\vec{d}_t$ for trailer $t$ based on its movement
        \STATE Extend direction vector $\vec{d}_t$ in the direction of movement
        \WHILE{the direction vector does not exit the image frame}
            \IF{the direction vector $\vec{d}_t$ intersects or overlaps with a vehicle object $v$}
                \STATE Associate trailer $t$ with vehicle $v$
                \STATE Break
            \ENDIF
        \ENDWHILE
    \ENDIF
\ENDFOR
\STATE \textbf{Return:} Associated trailer-vehicle pairs
\end{algorithmic}
\end{algorithm}

\subsection{Tire Tracker for Axle Counting}
\subsubsection{\textbf{Tire Tracker:}}
Vehicle tracking is needed to track the identity of a vehicle across video frames and differentiate with other vehicles in the scene. Similarly, tire tracking is required to identify the same tires across multiple frames. When the vehicle and the tires are clearly visible in a single frame, axle counting is straightforward. However, this is not always the case, our TRAX Algorithm \ref{algo:tire_tracker} tracks tires in such complex scenarios. Particularly,
\begin{itemize}
    \item The vehicle can be too long to fit in the field of view of the camera. Refer to Fig. ~\ref{fig:hard-example}.
    \item All tires may not be clearly visible in any single frame. For e.g., rear tires cannot be seen clearly until they come close to the camera, and by that time the front tires have moved out the frame.
\end{itemize}

\begin{algorithm}
\caption{TRAX : Tire Tracker for Axle Counting}
\label{algo:tire_tracker}
\begin{algorithmic}[1]
\STATE \textbf{Input:} $(x, y, t)$ data of tires of a single vehicle
\STATE Project $(x, y, t)$ onto motion direction using PCA. \\$ z_i = \mathbf{v}^\top (\mathbf{x}_i - \bar{\mathbf{x}}), 
\mathbf{x}_i = \begin{bmatrix} x_i & y_i \end{bmatrix}, \mathbf{v} : eigenvector
$
\STATE Apply inverse transform: $z=1/(1+c*z)$
\STATE Estimate the average slope of tracks in $(z, t)$ space
\WHILE{Tracks exist}
    \STATE Extract one track
    \STATE Store extracted track
\ENDWHILE
\STATE Pick a starting point $(t, z)$ for extracting a new track
\WHILE{New track is being formed}
    \STATE Find the candidate point $z_{next}$ at $(t+1, z)$ for extending the track
    \IF{$z_{next}$ exists}
        \STATE Extend the track by appending $z_{next}$
        \STATE $t \leftarrow t + 1$, $z \leftarrow z_{next}$
        \STATE Delete the points belonging to the current track
    \ELSE
        \STATE Store the current track
    \ENDIF
\ENDWHILE
\end{algorithmic}
\end{algorithm}

\subsubsection{\textbf{Axle Counting}}
As detailed in Algorithm \ref{algo:tire_tracker} The axle counting is carried out by a 2D projection based tire tracker. By converting spatiotemporal information to motion aligned coordinates, the algorithm supports effective tire track extraction and analysis.
We first gather spatiotemporal coordinates of tire positions as points $(x, y, t)$ where $x$ and $y$ correspond to spatial coordinates and $t$ is a time index. Project the $(x,y,t)$ on the direction of movement of vehicle to obtain a representation in terms of the function $z=T(x,y,t)$ using PCA\cite{ref15}. This procedure aligns the information along the primary axis of movement and simplifies analysis later.
Carry out an inverse transformation to rescale the time axis by using the equation $z=1/(1+c*z)$, where $\alpha$ is the scaling factor. This adjustment corrects differences in vehicle speeds or sampling rates to achieve uniform time spacing. Make an estimate of the slope of the tire tracks in the transformed space of $(z,t)$. This slope gives information about vehicle velocity and direction as time proceeds. Create new tracks by taking any starting point with given time $t$ and position $z$ coordinates.
Strive to find successive position $z_{next}$ at successive time $t+1$ corresponding to the existing position.
Add $z_{next}$ to the existing track and set current time to $t+1$ and position to $z_{next}$, where such a point is present. Remove the presently added point from database so that it is not repeated.
If we find no appropriate $z_{next}$, we finalize this existing track and save it to analyze later.
The final axle count refers to the overall tire tracks calculated for every given vehicle.

\section{Experiments and Results}

\subsection{Data}

\subsubsection{\textbf{Training Dataset}}
For training the vehicle detector and tire detection models we collected and annotated 1200 videos from the cameras installed on toll booths, the annotation process is detailed below. These 1200 videos are collected from 40 cameras, then extracted into 80000 frames. We split the data into 90\% images for training and 10\% for validation. The training data contains 147639 vehicles and 193217 tires. The validation data consist of 14211 vehicles and 20334 tires. \\

The detailed data distribution of Vehicles and Tire classes is given in Table \ref{tab:train_data_distribution}.

{
\begin{table}[t]
\centering
\caption{Label Distribution in Training and Validation Sets}
\begin{tabular}{|l|r|r|}
\hline
\textbf{Label Name} & \textbf{\#Train Labels} & \textbf{\#Val Labels} \\
\hline
Tire          & 193,217 & 20,334 \\
SUV           & 43,625  & 3,501  \\
Sedan         & 32,859  & 3,685  \\
Pickup\_Truck & 36,677  & 3,426  \\
Truck         & 8,403   & 779    \\
Semi\_truck   & 12,593  & 1,499  \\
Van           & 4,387   & 492    \\
Trailer       & 6,610   & 394    \\
Hatchback     & 2,099   & 339    \\
Bus           & 386     & 96     \\
\hline
\textbf{Total} & \textbf{340,856} & \textbf{34,545} \\
\hline
\end{tabular}
\label{tab:train_data_distribution}
\end{table}
}

\subsubsection{\textbf{Data Annotation Process:}}
The annotation of vehicle and tire detector models involved a multi-stage pipeline structured to produce high-quality labeled data efficiently by integrating automated segmentation with refinement done by humans. The annotation started off with the acquisition of video data from multiple sources, capturing mixed scenes of vehicles and road scenes. The videos were then preprocessed by extracting individual frames, basically converting the temporal video data to a dataset of many individual images that are easier to input to object detection tasks.

To minimize manual-labeling effort, the Segment Anything Model\cite{ref14} (SAMv2) was used to produce initial segmentation-based labels. SAMv2\cite{ref14}, a segmentation model driven by text prompts, was utilized with text prompts to detect relevant objects. The model was prompted with the text "vehicle" to detect vehicles and generated pixel-level segmentation masks corresponding to every vehicle in the frame. The masks indicating precise outline of every vehicle were further processed. Every segmentation mask was mapped to an enclosing polygon that captures outer boundary of vehicle region. The polygons were transformed to Oriented Bounding Boxes (OBBs) by performing geometric fitting (e.g., minimum-area rectangles). The reason vehicles were represented by OBBs is that they can be more flexible and can describe objects that appear at an angle or rotated since they are not limited to axis alignment. The same strategy was employed for tire detection. The SAMv2\cite{ref14} model was given the input word "tire" and generated segmentation masks only for tire areas in the image. Tires, however, were annotated in simpler Axis-Aligned Bounding Boxes (AABB) rather than oriented boxes. The boxes were obtained from the segmentation masks and represented in terms of (x,y,width,height) where (x,y) is the top-left corner of the box and width and height represent its size.

Following these automatically generated annotations, a manual correcting and validation process was performed to improve annotation accuracy. Each frame was inspected by human annotators to detect and correct mistakes, if any. This involved inserting missed detections (false negatives) where SAMv2\cite{ref14} didn't segment an object, deleting false positives where irrelevant areas were mistakenly marked, and adjusting bounding box sizes or orientations as needed to refine precision. This mixing of automated segmentation and manual refinement greatly smoothed out the annotation process while keeping accuracy high, finally delivering a well-annotated dataset that can be used to train vehicle and tire detection models.

\subsubsection{\textbf{Test Dataset}}
Based on the complexity we curated 3 test datasets.
\begin{enumerate}
    \item $Easy$ - Vehicles with only 2 axles including Sedans, SUVs, Vans and Hatchbacks. As shown in Fig. \ref{fig:easy_medium_example} left

    \item $Medium$ - Long Vehicles having 2 or more axles and all the tires are visible in frame including pickup trucks, trucks, semi trucks, trailers and buses. As shown in Fig. \ref{fig:easy_medium_example} right.

    \item $Hard$ -  Long vehicles having 2 or more axles but all the tires are not visible in any of the video frame. As shown in Fig. \ref{fig:hard-example}
    
    \begin{figure}[b] 
    \centering
    \includegraphics[width=0.85\linewidth,keepaspectratio]{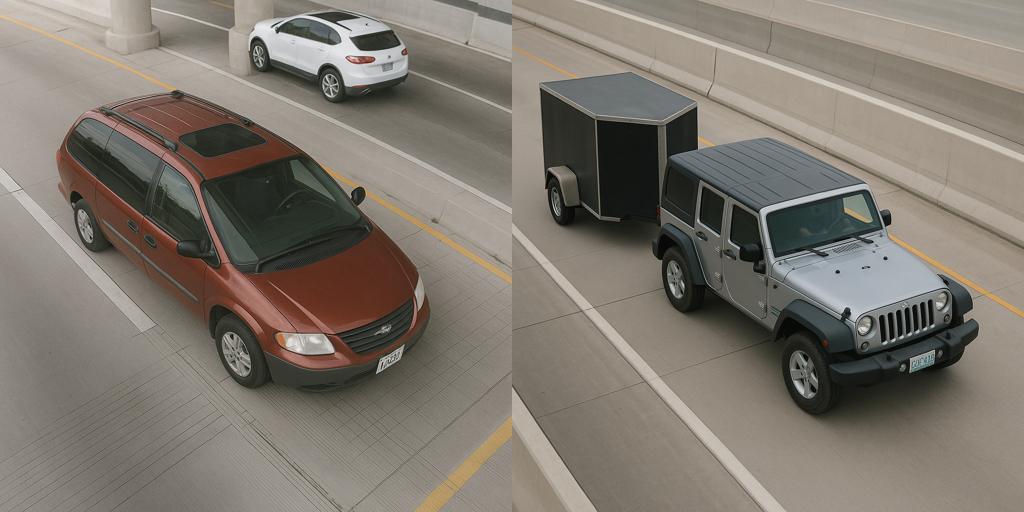} 
    \caption{Easy : Sedan, Medium : Trailer}
    \label{fig:easy_medium_example} 
    \end{figure}

    \begin{figure}[b] 
    \centering
    \includegraphics[width=0.85\linewidth,keepaspectratio]{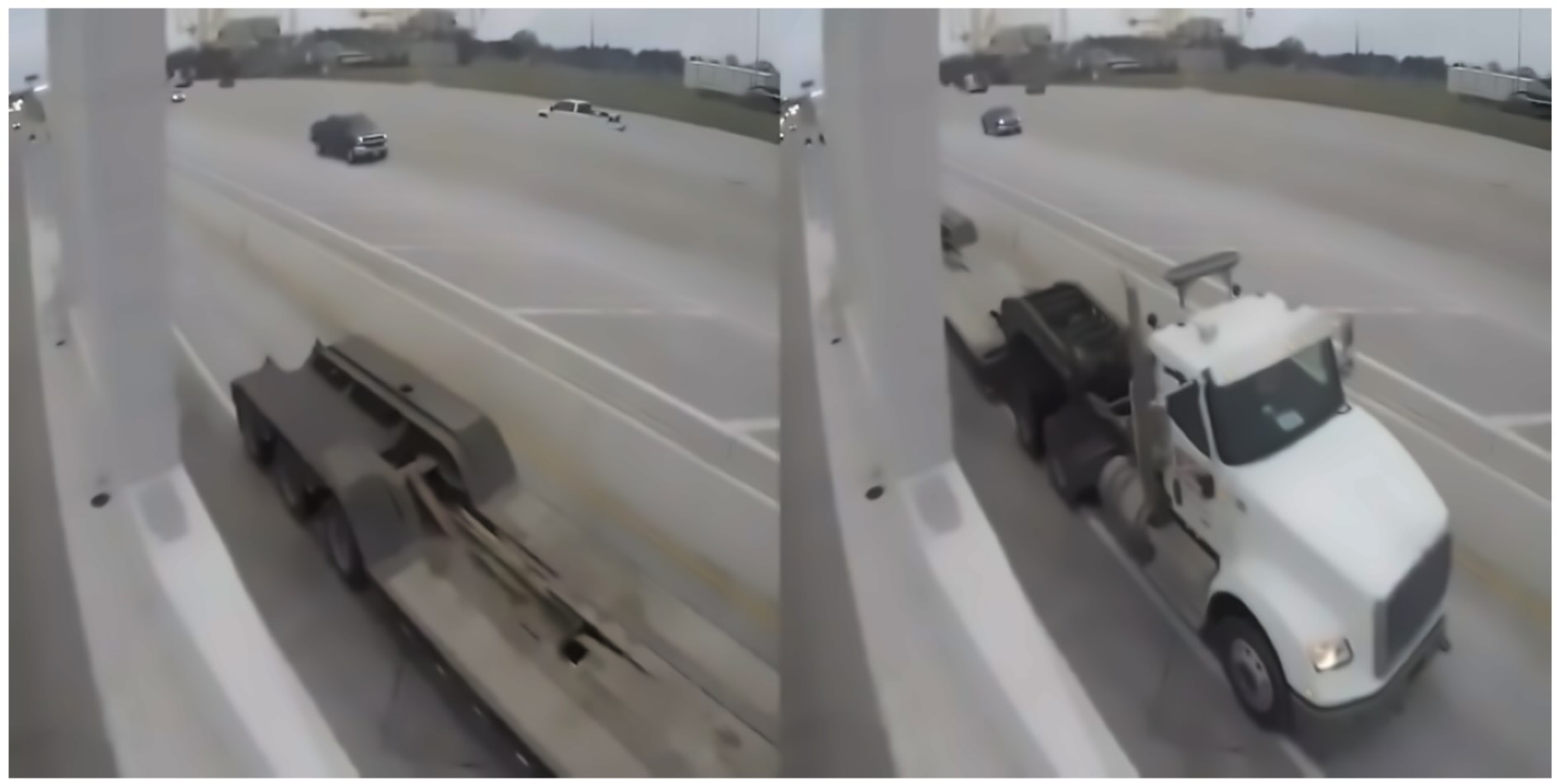} 
    \caption{Hard : Long vehicle}
    \label{fig:hard-example} 
    \end{figure}
    
\end{enumerate}
The test data set distribution is given in Table \ref{tab:test_data}


{
\begin{table}[t]
\centering
\caption{Summary of test data and corresponding labels.}
\begin{tabular}{|l|l|c|c|c|}
\hline
\textbf{Dataset} & \textbf{\# Images} & \textbf{\#Vehicle Labels} & \textbf{\#Tire Labels} \\
\hline
$Easy$ & 5455 & 19771 & 23141 \\
$Medium$ & 1624 & 3112 & 4159 \\
$Hard$ & 2091 & 4106 & 5487 \\
\hline
\end{tabular}
\label{tab:test_data}
\end{table}
}

\subsection{Experiments and Implementation Details}
As mentioned in \ref{sebsec:vehicle_detector} and Souza et al. (2024) \cite{ref2} work we picked YOLOv5 as our choice of detector model and Oriented Bounding Boxes over Axis Aligned Bounding Boxes.\\ 
We use mAP@50 score to evaluate Vehicle and Tire Detection models and Accuracy for Axle count evaluation. Also, Training independent models for Tire and Vehicle detection yields 4\% better mAP on Tire Class. The results of Vehicle detection, Tire detection and Axle count accuracy with and without TRAX are given in Table \ref{tab:test_set_metrics}
We have compared TRAX accuracy agains traditional Li et al. (2021) \cite{ref3}'s, frequency counting based axle counting method.\\
\textbf{Frequency based axle counting:} First we count number of tires associated to each vehicle (tracked by vechile tracker) on every frame of the video. Then the count with the maximum frequency is chosen as the final axle count.

\begin{table}[t]
\centering
\caption{Test Set Results.}
\label{tab:test_set_metrics}
\begin{tabular}{|l|c|c|c|c|}
\hline
\textbf{Test Set} & \textbf{Vehicle mAP@50} & \textbf{Tire mAP@50} & \textbf{Mode} & \textbf{TRAX} \\
\hline
$Easy$   & 0.750 & 0.988 & 0.960 & \textbf{0.998} \\
$Medium$ & 0.740 & 0.930 & 0.7 & \textbf{0.920} \\
$Hard$   & 0.648 & 0.810 & 0.548 & \textbf{0.928} \\
\hline
\end{tabular}
\end{table}

\subsubsection{\textbf{Vehicle and Tire Detector Training}}

The Yolov5s-obb and Yolov5s is trained with CrossEntropy loss and Object detection loss as used in YOLOv5\cite{ref8} for multiclass vehicle detection and tire detection respectively. The optimizer used is SGD with momentum 9.4e-1 and weight decay 5e-3 for 300 epochs. We augment the training data with flipLR, flipUD, Mosiac and mixup augmentation during training, the images are resized to 320x320 input resolution. The initial learning rate is initialized with 1e-2 and then reduced by the factor of 0.2 using One Cycle LR schedular\cite{ref12} LR decay. The last epoch checkpoint is used as the final checkpoint.

\subsubsection{\textbf{Vehicle Tracker}}
The choice of YOLOv5\cite{ref8} is based on the work done by \cite{ref2} Souza et al. (2024) which tested a number of different YOLO\cite{ref7} variants (v5, v6, v7, v8) and discovered that the best precision (99.40) and recall (98.20) was produced by YOLOv5m\cite{ref8} for detecting tires.\\
Vehicle Tracker is based on DeepSORT\cite{ref5}, DeepSORT uses a visual feature extractor to match objects. For Visual feature extractor we have trained OSNet\cite{ref6}, on 100 randomly sampled vehicles for training and 20 for testing. The OSNet\cite{ref6} is trained using default setting mentioned in Torch ReID repo\cite{ref10}. Table \ref{tab:osnet_results} show the evaluation results.


{
\begin{table}[t]
\centering
\caption{OSNet Results.}
\label{tab:osnet_results}
\begin{tabular}{|l|c|c|c|c|c|}
\hline
\textbf{Metric} & \textbf{Rank-1} & \textbf{Rank-5} & \textbf{Rank-10} & \textbf{Rank-20} & \textbf{mAP} \\
\hline
CMC Curve & 92.0 & 96.0 & 97.3 & 98.2 & 93.8 \\
\hline
\end{tabular}
\end{table}
}

With this trained OSNet\cite{ref6} feature extractor evaluated with CMC curve \cite{ref11}, the DeepSORT\cite{ref5} tracker is independently evaluated for tracking accuracy using Multi Object Tracking Accuracy \textbf{MOTA}\cite{ref13} metric. The details are provided in Table \ref{tab:deepsort_results}


{
\begin{table}[htbp]
\centering
\caption{DeepSORT Results.}
\label{tab:deepsort_results}
\begin{tabular}{|l|c|}
\hline
\textbf{Test Set} & \textbf{MOTA} \\
\hline
$Easy$ & 96.1 \\
$Medium$ & 88.3 \\
$Hard$ & 88.1 \\
\hline
\end{tabular}
\end{table}
}

\subsubsection{\textbf{TRAX : Tire Tracker for Axle Count}}
\begin{figure}[htbp] 
    \centering
    \includegraphics[width=0.85\linewidth,keepaspectratio]{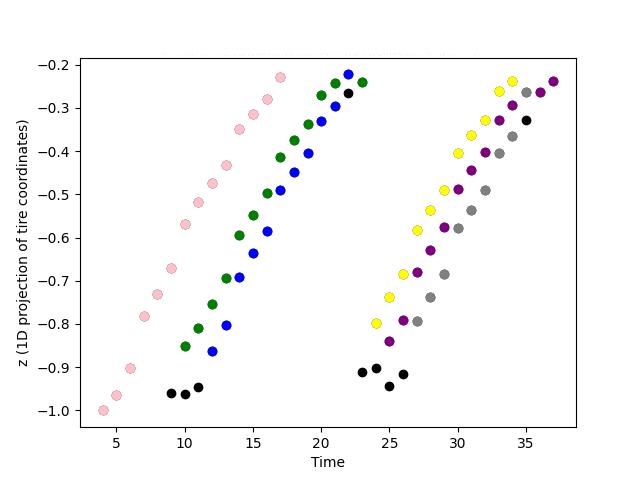} 
    \caption{2D projection of 6 Axle Long vehicle}
    \label{fig:axle-example} 
\end{figure}

The TRAX algorithm is detailed in {Algorithm} \ref{algo:tire_tracker}. An example of 2D projection of $(x,y,t)$ values of long vehicle having 6 axle counts is given in Fig \ref{fig:axle-example}. Note that different axles are color coded in different colors, black color are the instances which are discarded by the algorithm. Here, X-axis represents the relative timestamp and Y-axis represents projection $z = T(x, y, t)$. In this example, the rear 3 axles appear in front of camera when front 3 axles are out of camera view. \\ 
The accuracy of TRAX is evaluated using the end to end accuracy of axle counts over the test datasets, details are provided in Table\ref{tab:test_set_metrics}

\section{Conclusion}
We introduce a novel tire-based axle counting algorithm called TRAX, carefully designed to tackle the challenge of axle counting in surveillance videos, where the appearance of all tires on a given axle cannot be guaranteed in every frame. Contrary to previous full visibility or simplistic heuristic based approaches for counting, TRAX tracks partial and occluded vehicle's tire detections sequentially through the frames and leverages learned motion behavior and temporal consistency to infer axle presence and count at high accuracy.

TRAX proves strong performance at all levels of scene complexity, with 99.8\% accuracy on easy testing and demonstrating outstanding results at 92\% and 92.8\% when using medium and hard datasets, respectively. The benchmarks feature a variety of real-world traffic environments with mixed vehicle speeds, types, and occlusion profiles.

The end-to-end pipeline is robust to unfavorable environmental scenarios like low-light visibility, rain, snow, and fog instances. This is due to the synergy between spatio-temporal reasoning of TRAX and pre-procssing enhancements applied to the input. However, we observe performance dips in certain edge cases, specifically when the structural profile of the vehicle is complex (e.g., car transporters) or when over 60\% of the vehicle or tire region is occluded by neighboring vehicles, rainwater accumulation in case of flood rain, or dense fog and night time darkness. These occlusions interfere with the performance of both vehicle and tire detection models, which are integral to the initial step of TRAX. Although these failures exist, we consider them as opportunities for further development. Overall, this work represents a major step forward in developing scalable, AI-driven axle counting systems, paving the way for machine vision to eventually replace traditional infrastructure.

\end{document}